\documentclass[letterpaper, 10pt, conference]{ieeeconf}
\IEEEoverridecommandlockouts	% to override locked commands

\usepackage{multirow}
\usepackage{lipsum} 
\usepackage{amsmath}
\usepackage{amssymb}
\usepackage{subcaption}
\usepackage{graphicx}
\usepackage{diagbox}
\graphicspath{{./Figures/}}
\usepackage{glossaries}
\usepackage[table,xcdraw]{xcolor}
\usepackage[ruled,vlined]{algorithm2e}
\usepackage[final]{listings}
\usepackage{longtable}
\usepackage[hidelinks]{hyperref}
\usepackage{tikz,pgfplots}

\newacronym{ITS}{ITS}{Intelligent Transportation Systems}
\newacronym{ROS}{ROS}{Robot Operating System}
\newacronym{URDF}{URDF}{Unified Robot Description Format}
\newacronym{SLAM}{SLAM}{Simultenous Localization and Mapping}
\newacronym{JSON}{JSON}{JavaScript Object Notation}
\newacronym{LDS}{LDS}{Laser Distance Sensor}
\newacronym{iCab}{iCab}{Intelligent Campus Automobile}
\newacronym{AV}{AV}{Autonomous Vehicle}
\newacronym{HMI}{HMI}{Human Machine Interface}
\newacronym{PAFs}{PAFs}{Part Affinity Fields}
\newacronym{VRUs}{VRUs}{Vulnerable Road Users}

\title{\LARGE \bf Autonomous Driving: Framework for Pedestrian Intention Estimation in a Real World Scenario}

\author{Walter Morales Alvarez$^{1}$ \emph{Student Member, IEEE}, Francisco Miguel Moreno$^{2}$, Oscar Sipele$^{1,3}$, Nikita Smirnov$^{1,4}$, \\
Cristina Olaverri-Monreal$^{1}$ \emph{Senior Member, IEEE}%
\thanks{$^1$ Johannes Kepler University Linz, Austria; Chair  Sustainable Transport Logistics 4.0. \texttt{\{walter.morales\_alvarez, b.\_oscar.sipele\_siale, nikita.smirnov, cristina.olaverri-monreal\}@jku.at}}%
\thanks{$^2$ Universidad Carlos III de Madrid, Spain; Intelligent Systems Lab. \texttt{franmore@ing.uc3m.es}}
\thanks{$^3$ Universidad Carlos III de Madrid, Spain; Computer Science Department. \texttt{bsipele@inf.uc3m.es}}
\thanks{$^4$ Ural Federal University, Department of Communications Technology.}
}

\newcommand\copyrighttext{%
  \footnotesize \textcopyright 2020 IEEE. Personal use of this material is permitted. Permission from IEEE must be obtained for all other uses, in any current or future   media, including reprinting/republishing this material for advertising or promotional purposes, creating new collective works, for resale or redistribution to servers or lists, or reuse of any copyrighted component of this work in other works.  DOI: \href{https://ieeexplore.ieee.org/abstract/document/9304624}{10.1109/IV47402.2020.9304624}}

\newcommand\copyrightnotice{%
\begin{tikzpicture}[remember picture,overlay]
\node[anchor=south,yshift=10pt] at (current page.south) {\fbox{\parbox{\dimexpr\textwidth-\fboxsep-\fboxrule\relax}{\copyrighttext}}};
\end{tikzpicture}%
}

\begin{document}

\maketitle
\copyrightnotice
\thispagestyle{empty}
\pagestyle{empty}

%%%%%%%%%%%%%%%%%%%%%%%%%%%%%%%%%%%%%%%%%%%%%%%%%%%%%%%%%%%%%%%%%%
\begin{abstract}
Rapid advancements in driver assistance technology will lead to the integration of fully autonomous vehicles on our roads that will interact with other road users. To address the problem that driverless vehicles make interaction through eye contact impossible, we describe a framework for estimating the crossing intentions of pedestrians in order to reduce the uncertainty that the lack of eye contact between road users creates.  
The framework was deployed in a real vehicle and tested with three experimental cases that showed a variety of communication messages to pedestrians in a shared space scenario. Results from the performed field tests showed the feasibility of the presented approach.
\end{abstract}

%%%%%%%%%%%%%%%%%%%%%%%%%%%%%%%%%%%%%%%%%%%%%%%%%%%%%%%%%%%%%%%%%%

\section{Introduction}
\label{sec:introduction}

Rapid advancements in driver assistance technology will lead to the integration of fully autonomous vehicles that do not need a driver in a variety of driving and pedestrian environments. 
In highway environments, autonomous vehicles need to take into account the state of the surrounding vehicles. However, in urban and semi-urban environments the perception complexity regarding other road users increases as vulnerable road users such as cyclists or pedestrians, who lack physical protection against collisions, also share the road with vehicles~\cite{olaverri-monreal2016shadow}.

In a traffic interaction where any combination of two or more vehicular units or road users encounter each other, each is obliged to take the others into account to avoid a potentially unsafe situation. In this interaction eye contact or its avoidance plays a central role as the integration of glances facilitates cooperative action.

In traffic that involves driverless vehicles, visual or audio messages might replace eye contact-based communication with other road users to make sure that the intentions of all road users are understood by all entities in the environment and the corresponding actions can be performed according to each case. 

Interaction between autonomous vehicles and pedestrians can be addressed by judging and anticipating the actions of the different actors in the system and determining the rules for their co-existence~\cite{allamehzadeh2016automatic} and by creating protocols that allow to develop a level of trust in autonomous vehicles equal to human driven vehicles \cite{hussein2016p2v}.

Thus, in this paper we develop a framework for estimating the intention of pedestrians using state of the art algorithms. We then use the framework to predict pedestrian  crossing behavior when they are exposed to a level 5 autonomous vehicle that integrates communication interfaces. 

The remainder of the paper is organized as follows: the next section describes related work in the field; section~\ref{sec:framework} explains the implemented framework to predict  pedestrian behavior.  Section~\ref{sec:fieldtest} presents  the  method used to  assess  the  data collected;  section~\ref{sec:results} presents  the  obtained  results;  and finally, section~\ref{sec:conclusion} discusses and concludes the work. 
%%%%%%%%%%%%%%%%%%%%%%%%%%%%%%%%%%%%%%%%%%%%%%%%%%%%%%%%%%%%%%%%%%
\section{Related Work}
\label{sec:relatedwork}

This section describes related literature regarding interaction between pedestrians and autonomous vehicles and the development of algorithms that determine pedestrian crossing intention.

In \cite{Habibovic2018} and \cite{Burns2019} the authors studied the impact of different communication interfaces on safety, their results showing an increase  in perceived safety  when the vehicle was equipped with a communication interface. 

In the same line of research, the authors of \cite{Matthews} studied different types of interfaces that indicated explicitly to pedestrians whether they could cross or not, obtaining a priority dependence between pedestrian behavior and their distance from the vehicle. 

More complex communication protocols have been implemented in further works \cite{Mahadevan2018}. For example, through simulated artificial eyes that follow the pedestrians \cite{Chang2017a} or through vehicle driving patterns \cite{Beggiato2018} like acceleration and deceleration. 

Furthermore, other studies not only address the impact of the type of communication interface on pedestrians, but also the time frame in which  the messages are displayed and the size of the vehicle involved \cite{DeClercq2019}, as well as the impact of interfaces on certain population groups such as children \cite{Charisi2017}. 

Although the previous studies establish an increase in trust in automation in a crossing situation, most of the results of these studies are based on qualitative data, simulations or Wizard of OZ paradigms, all of which can have an impact on the study, and they are not based on real situations, research on interaction between pedestrians and AV in real situations being very limited \cite{8813899}.

Pedestrian tracking algorithms that generate sequential representations of pedestrians and with these classify pedestrians' actions have been presented in \cite{Ludl2019}, \cite{Zhan2019}. The authors used convolutional neural networks (CNN) based on spatial parameters to determine whether or not a given pedestrian would cross the road.

Other works use recurrent neural networks (RNN) to model sequential data and based on these they classify pedestrian actions \cite{Veeriah}, \cite{Mahmud2017}, \cite{Donahue2015} or their crossing intention \cite{rasouli2017they}. 

Although all studies present relevant results for estimating the action and intention of pedestrians, they either lack of cross validation in real environments or only focus on the development of models whose inputs come from labeled datasets like JAAD dataset or PIE dataset \cite{rasouli2017pie} \color{black}. Therefore, several processing steps must be performed to allow the models to be used online.

Thus the main contributions of the work presented in this paper are:

\begin{itemize}
	\item The development of a framework for predicting pedestrians crossing intention using state of the art algorithms, such that using only the data acquired through the vehicle sensors it is possible to predict pedestrian intention.  
	\item  The application of the developed prediction framework to perform a empirical comparison between three different cases in a real-world shared space scenario in which pedestrians and autonomous vehicles interact. 
\end{itemize}

%%%%%%%%%%%%%%%%%%%%%%%%%%%%%%%%%%%%%%%%%%%%%%%%%%%%%%%%%%%%%%%%%%

\section{Framework}
\label{sec:framework}

We relied on the algorithms use in \cite{rasouli2017they,Morales-Alvarez2018,Alvarez2019,cao2018openpose,marin2016stereo} to develop the proposed framework to estimate pedestrian crossing intention in the presence of an autonomous vehicle.

\begin{figure*}[!ht]
	\centering
	\includegraphics[width=0.95\textwidth]{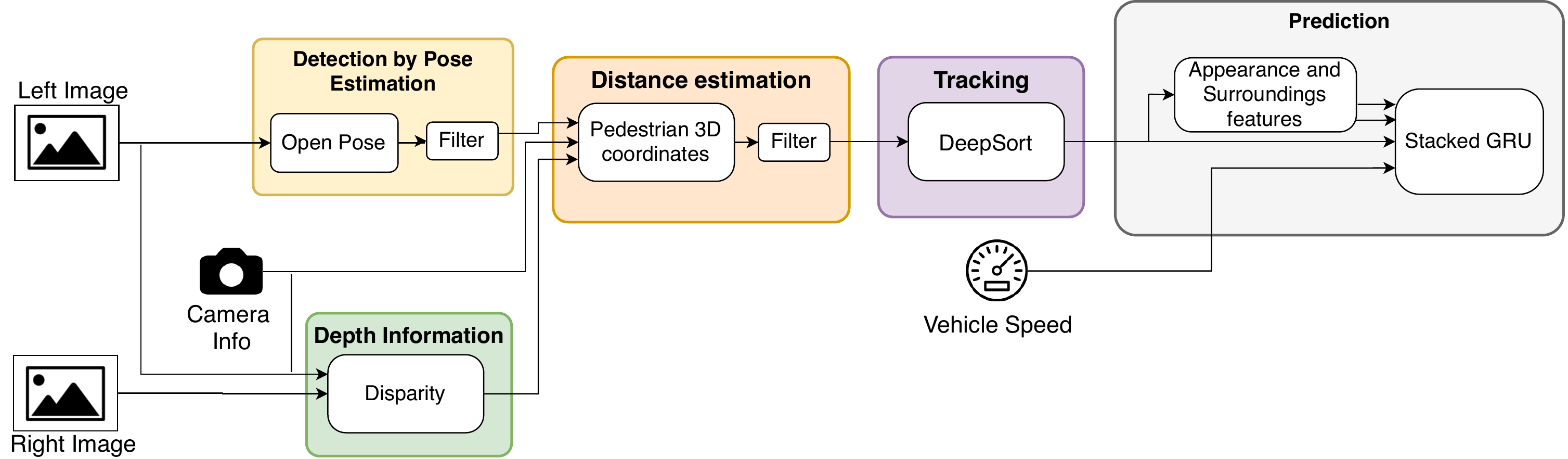}
	\caption{Architecture of the analyzing algorithm, inputs being Left Image, Right Image, intrinsic information of stereo camera and vehicle speed. Each module corresponds to a different algorithm used to estimate a particular pedestrian's intention. }
	\label{fig:architecture}
\end{figure*}

The framework uses the data acquired by the sensors of an autonomous vehicle to extract pedestrian positions in real coordinates, their poses, their local context and the speed of the vehicle. Using this information it estimates pedestrian intention in terms of whether they will cross or not. The main architecture of the algorithm is shown in Figure~\ref{fig:architecture}.

\subsection{Pedestrian detection and pose estimation}
Initially, the framework detects pedestrians and estimates their poses in each image acquired by vehicle's camera. This process is done using OpenPose library developed by the Carnegie Mellon University Perceptual Computing Lab in \cite{cao2018openpose},\cite{cao2017realtime},\cite{wei2016cpm}. With this library we obtained 25 pedestrian pose keypoints that are used to extract a pedestrian's bounding box in the image of the camera, as depicted in \cite{Morales-Alvarez2018}. These detections and poses will be used in the intention predicting model as input to estimate a pedestrian’s  crossing action.

Although OpenPose estimates the poses of pedestrians, the library is not perfect and in some cases it erroneously estimates  poses in places where there are no pedestrians, computes poses that are proportionally unlikely to belong to a human or obtains poses with too few pose keypoints. Therefore, the poses estimated by the library are filtered, discarding those poses with less than 20 pose keypoints or those in which the width of the bounding box is greater than the height.

\subsection{Pedestrian distance estimation}

The estimation of the distance between the autonomous vehicle and pedestrians is made using the depth information extracted through the stereo camera that is present in the autonomous vehicle. This depth information is determined by the framework calculating the disparity between the left and right images acquired by the stereo camera. The disparity is calculated using the block matching algorithm as in \cite{marin2016stereo}.

Initially, the stereo camera was calibrated to determine its intrinsic parameters and to be able to rectify the acquired images, in order to apply the block matching and distance estimation algorithm effectively. Explicitly, the calibration of the camera allowed us to obtain the focal length ($f$), the base pixels ($c_x'$,$c_y'$) and the distance between the individual cameras ($B$) of the stereo camera. Having the disparity and the intrinsic parameters of the camera, the 3D coordinates can be calculated using the equations presented in \cite{Morales-Alvarez2018}.

During this process, the framework filters out those pedestrians whose distance to the vehicle is greater than 15 meters. At this distance, the pose keypoints obtained presented an error due to the scale of pedestrians. Also, for this use case, pedestrians outside this range were not of interest since the vehicle was not an impediment to crossing as for safety reasons the vehicle did not exceed 20 km/h. 

\subsection{Pedestrian Tracking}

In order to predict pedestrian intention, a sequential representation of each individual must be obtained by tracking their movement along several images. To achieve this we implemented the DeepSort algorithm \cite{Wojke2018a}, \cite{Wojke}, which is an extension of Simple Online and Realtime Tracking (SORT) \cite{Bewley2016}. 

Deepsort is an algorithm that matches pedestrian features extracted in a frame with features extracted in previous frames. For this purpose, DeepSort calculates a motion matching degree and an apparent matching degree to establish the correspondence between pedestrian detections in previous frames and the detections in the current one. The motion matching degree is calculated using a Kalman filter, which predicts the location of a pedestrian’s bounding box in the next frame. In this way, the motion matching degree is obtained by calculating the Mahalanobis square distance between the Kalman filter prediction and a pedestrian’s detection in the current frame.

To calculate the apparent matching degree, the authors of DeepSort propose the computation of an appearance descriptor $\textbf{r}_j$ with $\lVert\textbf{r}_j\rVert = 1$ for each detection $\textbf{d}_j$ in the current frame, followed by the calculation of the cosine of the angle between these descriptors and the 100 descriptors saved by the tracker that correspond to pedestrian detections in previous frames. To compute these $\textbf{r}_j$ descriptors we opted for using the convolutional neural network (CNN) with the pre-trained weights on a re-identification dataset given by the authors of DeepSort.
Finally, having the motion matching degree $d^{(1)}(i,j)$ and the apparent matching degree $d^{(2)}(i,j)$, the association problem 
between the previous track ($i$) and the current detections ($j$)
is solved by combining both parameters in a weighted sum.

\begin{equation}
c_{i,j} = \lambda d^{(1)}(i,j) + (1-\lambda) d^{(2)}(i,j)
\end{equation} 

where $c_{i,j}$ is the metric proposed in DeepSort, $\lambda$ a hyperparameter to associate both the motion degree and apparent degree.  This metric is used as input in the Hungarian algorithm to establish the final match between the previous tracks and the current detections. In our case, the detections used by DeepSort come from those obtained by OpenPose, unlike the original work of the developers of DeepSort who use a custom neuronal network to detect pedestrians. Using OpenPose detections allows us to keep pedestrians identified with their current pose.

\subsection{Pedestrian intention estimation}

To make predictions of the pedestrians’ intentions, we based our methods on those presented in \cite{rasouli2017they}. In that work, the authors designed a recurrent neural network (RNN) that takes into account the context of pedestrians observed in the past to predict whether each current one will cross or not as a binary classification task. To this end, their prediction relies on five sources of information including the local context of the given pedestrian (features of the pedestrian and their surroundings), then pedestrian's pose, their location, and the speed of the vehicle itself.

The model developed is based on a stacked RNN architecture, in which the features at each level are gradually merged depending on their complexity, leaving the visual features for the bottom layers and the dynamic features such as trajectory and speed at the highest levels. This stacked RNN uses Gated Recurrent Units (GRU) to evaluate the sequential data.

To implement this model we opted for using the pretrained weights on the PIE \cite{rasouli2017pie} dataset, and cross-validate it in a shared space scenario with the videos gathered in the performed field tests. \color{black}

In our framework we use the results of the previous modules as inputs of the stacked RNN as follows: 

\begin{itemize}
	\item\textbf{Local context:} The appearance and surroundings of each pedestrian was used. We define appearance as the features of a pedestrian’s image in each frame. The features of the pedestrian's surroundings were computed by extracting a square region of interest around the pedestrian that is proportional to the size of their bounding box. The pedestrian image was covered  by setting gray pixels of RGB value (128,128, 128) in their bounding box. Like the developers of the model, we resized the images to 224 $\times$ 224 and used a VGG16~\cite{Simonyan2015} model pretrained with Imagenet~\cite{Russakovsky2015}, followed by average pooling. 
	
\item\textbf{\textbf{Pose:}} Although we use the pose obtained by OpenPose, we discarded 7 keypoints from each post because the model was pretrained weights supports only 18 keypoints. We then normalize and concatenate the remaining keypoints to obtain a 36D vector feature.
	
	\item\textbf{\textbf{Pedestrian's Location:}} We calculated the relative displacement from the initial position of the bounding box of each pedestrian.
	
	\item\textbf{Speed:} We obtained the speed of the vehicle through its CAN bus that transmits the velocity of the vehicle in each instant of time.
\end{itemize}

Following the results of the developers of the model, the framework is responsible for predicting the intention of those pedestrians who are one second away from interacting with the autonomous vehicle, performing a tracking of pedestrians for 1.5 seconds prior to the moment of prediction.

%%%%%%%%%%%%%%%%%%%%%%%%%%%%%%%%%%%%%%%%%%%%%%%%%%%%%%%%%%%%%%%%%%
\section{Field Test Description}
\label{sec:fieldtest}

\begin{figure}
	\centering
	\includegraphics[width=0.35\textwidth]{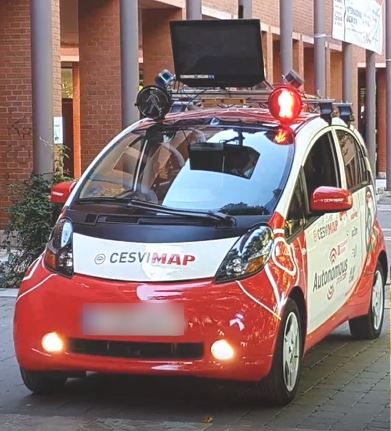}
	\caption{Autonomous vehicle used to perform the  field tests}
	\label{fig:apparatus}
\end{figure}

In order to be able to evaluate the framework and at the same time determine pedestrian behavior in front of an autonomous vehicle, we carried out a series of tests  using the Autonomous Driving Automobile (ADA) developed by the University Carlos III of Madrid. This is a Robotic Operating System (ROS)-based autonomous vehicle, equipped with perception sensors and control systems that allow the vehicle to drive without a human driver. Additionally, we installed in this vehicle a screen and a traffic light on top of the car that function as communication interfaces with surrounding pedestrians (see Figure~\ref{fig:apparatus}). These interfaces independently displayed a message to pedestrians indicating whether or not they had been detected by the autonomous vehicle. This message was developed as a C++ ROS node that receives a message from the vehicle's obstacle detection process that determines if it must to yield to pedestrians or not~\cite{demiguel2019}. Each interface was tested independently to determine if they affected pedestrian crossing behavior. Therefore we had the following test conditions:

\begin{itemize}
    \item \textbf{Baseline}: No communication interface was activated.
    \item \textbf{Screen display}: The screen displayed a pair of open eyes indicating pedestrian's that they  had  been  detected/could cross, or a pair of closed eyes indicating that the AV had not noticed the pedestrian \cite{Alvarez2019}.
    \item \textbf{Traffic light}: The installed traffic lights indicate pedestrians whether they can cross (green light) or not (red light).
\end{itemize}

For the tests, the vehicle drove autonomously for two days across the university campus to an intersection that is frequently used by pedestrians from the locality, as it connects two main local streets. To ensure pedestrians safety, a person seated in the backseat of the vehicle monitored the environment and stopped the car in critical situations (e.g distracted pedestrians). 
The vehicle's camera recorded the interaction between pedestrians and the autonomous vehicle. The recorded data was used as offline dataset to validate the framework presented in section~\ref{sec:framework}. A total of 15 videos of approximately 7 minutes each were obtained with 392 pedestrians.
Considering that the purpose of this work was to implement a framework that could be used as a system to predict pedestrian crossing intention when  exposed to an autonomous vehicle, we first needed to analyze and describe actual  pedestrian behavior  in order to compare it with the predicted results from the implemented framework. To this end, we relied on the pedestrian tracking algorithms previously described in section~\ref{sec:framework}, which we  applied throughout the different videos. We then performed a labeling process to the unlabeled data samples, including data that pertained to pedestrians that were walking beside the vehicle, extending thus the approach presented in \cite{Alvarez2019}. 

\begin{figure}
	\centering
	\begin{subfigure}{0.4\textwidth}
		\includegraphics[width=\textwidth]{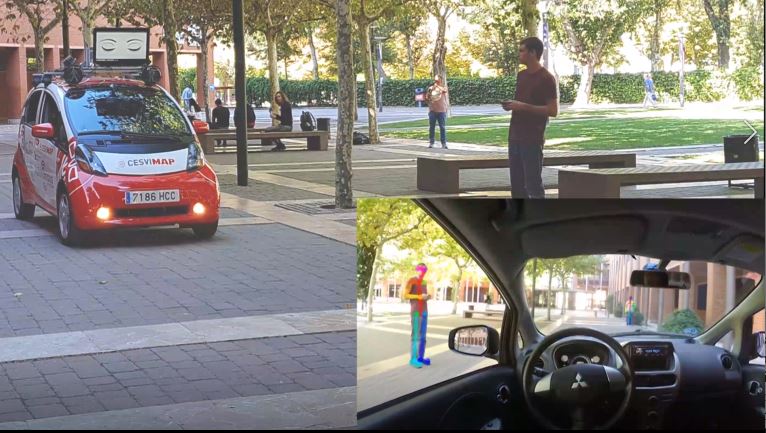}
		\caption{ }
	\end{subfigure}
	~ %add desired spacing between images, e. g. ~, \quad, \qquad, \hfill etc. 
	%(or a blank line to force the subfigure onto a new line)
	\begin{subfigure}{0.4\textwidth}
		\includegraphics[width=\textwidth]{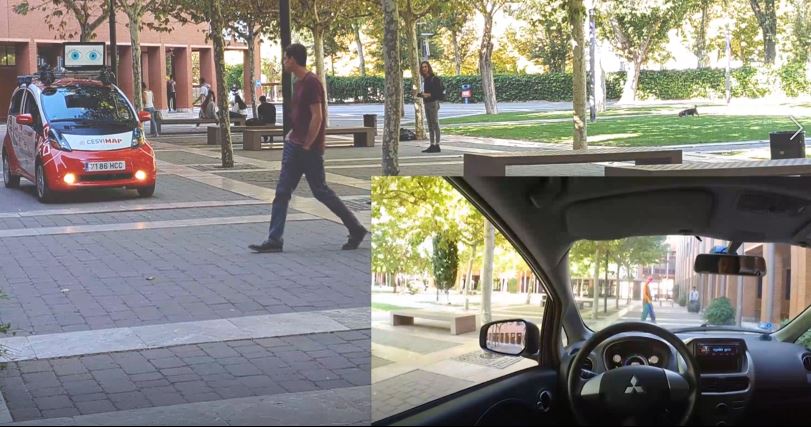}
		\caption{ }
	\end{subfigure}
	\caption{Example crossing situation where (a) corresponds to the vehicle displaying the closed eyes image to pedestrians and (b) the opened eyes image}\label{fig:vehicle}
\end{figure}

%%%%%%%%%%%%%%%%%%%%%%%%%%%%%%%%%%%%%%%%%%%%%%%%%%%%%%%%%%%%%%%%%%
\section{Results}
\label{sec:results}

After applying the implemented framework, a correct pedestrian detection of 93.21\%  was obtained, while  the illumination of the camera or the pedestrian density to which the algorithm was exposed caused problems in the remaining 6.79\% of cases.
Of the detected pedestrians, 87.04\% were successfully tracked, and the framework identified in successive frames a unique numerical id for each pedestrian.

\begin{figure}
	\centering
	\begin{subfigure}{0.45\textwidth}
		\includegraphics[width=\textwidth,scale=0.1]{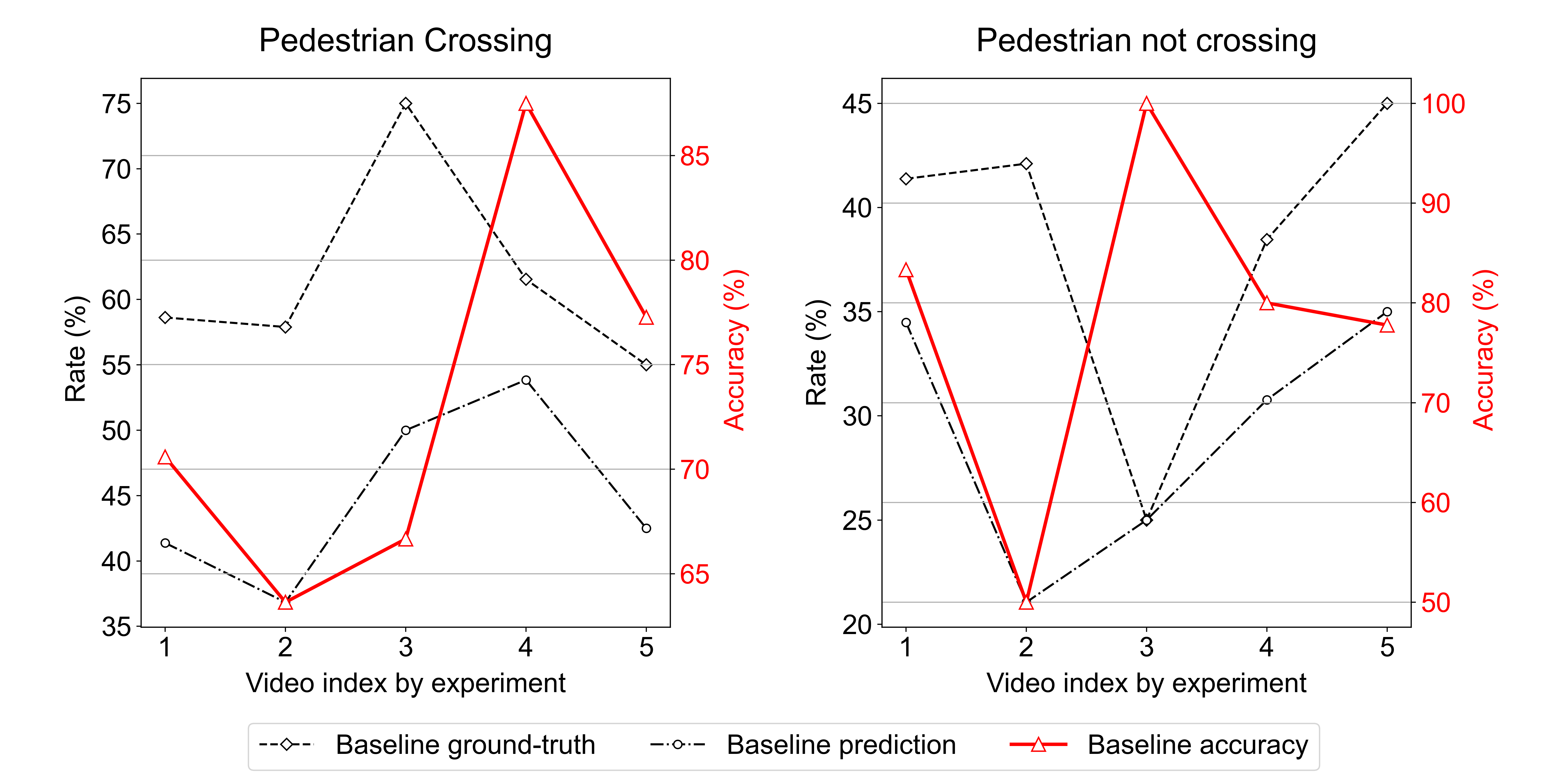}
		\caption{ }
	\end{subfigure}
	~ %add desired spacing between images, e. g. ~, \quad, \qquad, \hfill etc. 
	%(or a blank line to force the subfigure onto a new line)
	\begin{subfigure}{0.45\textwidth}
		\includegraphics[width=\textwidth]{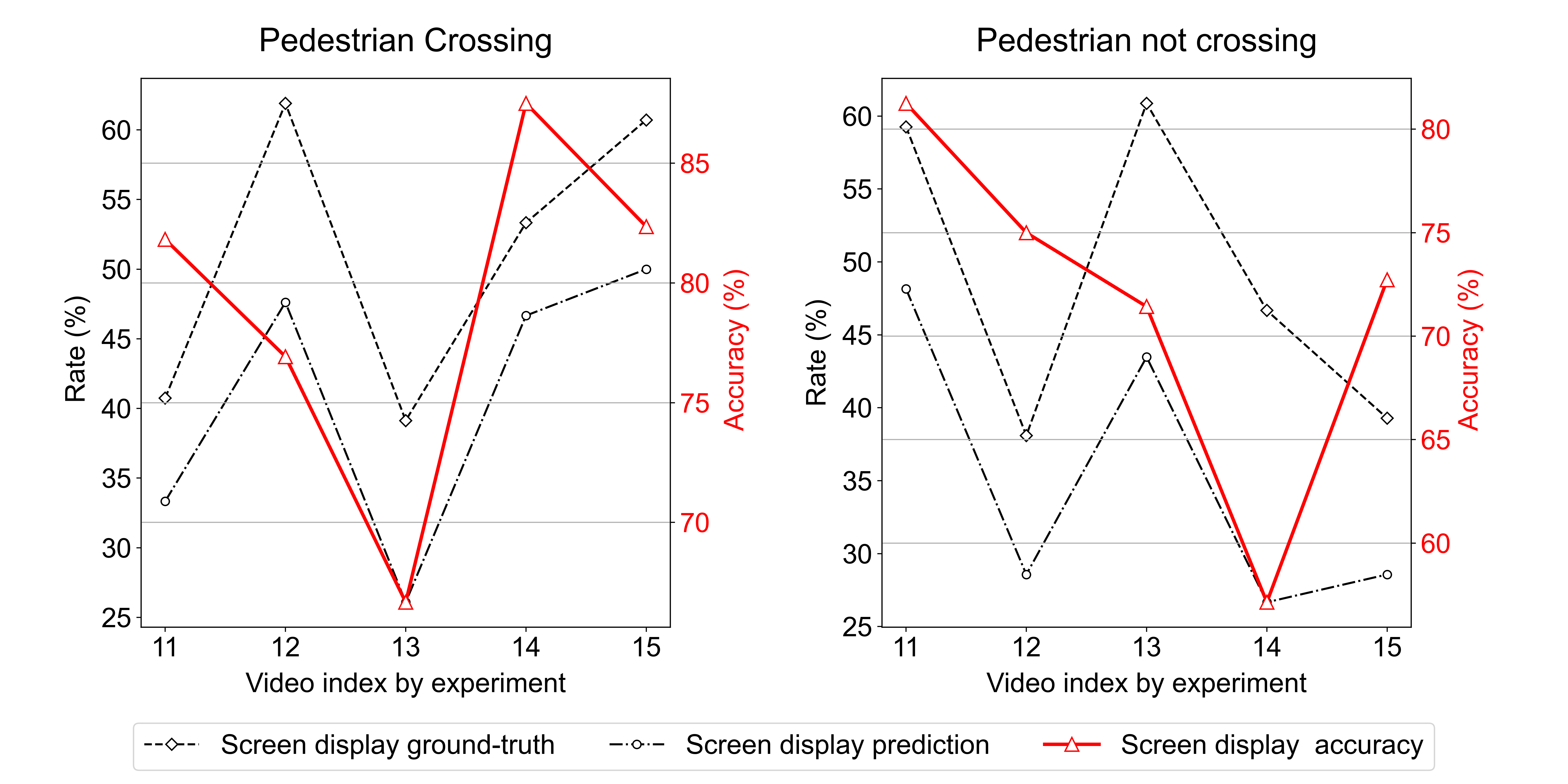}
		\caption{ }
	\end{subfigure}
	\begin{subfigure}{0.45\textwidth}
		\includegraphics[width=\textwidth]{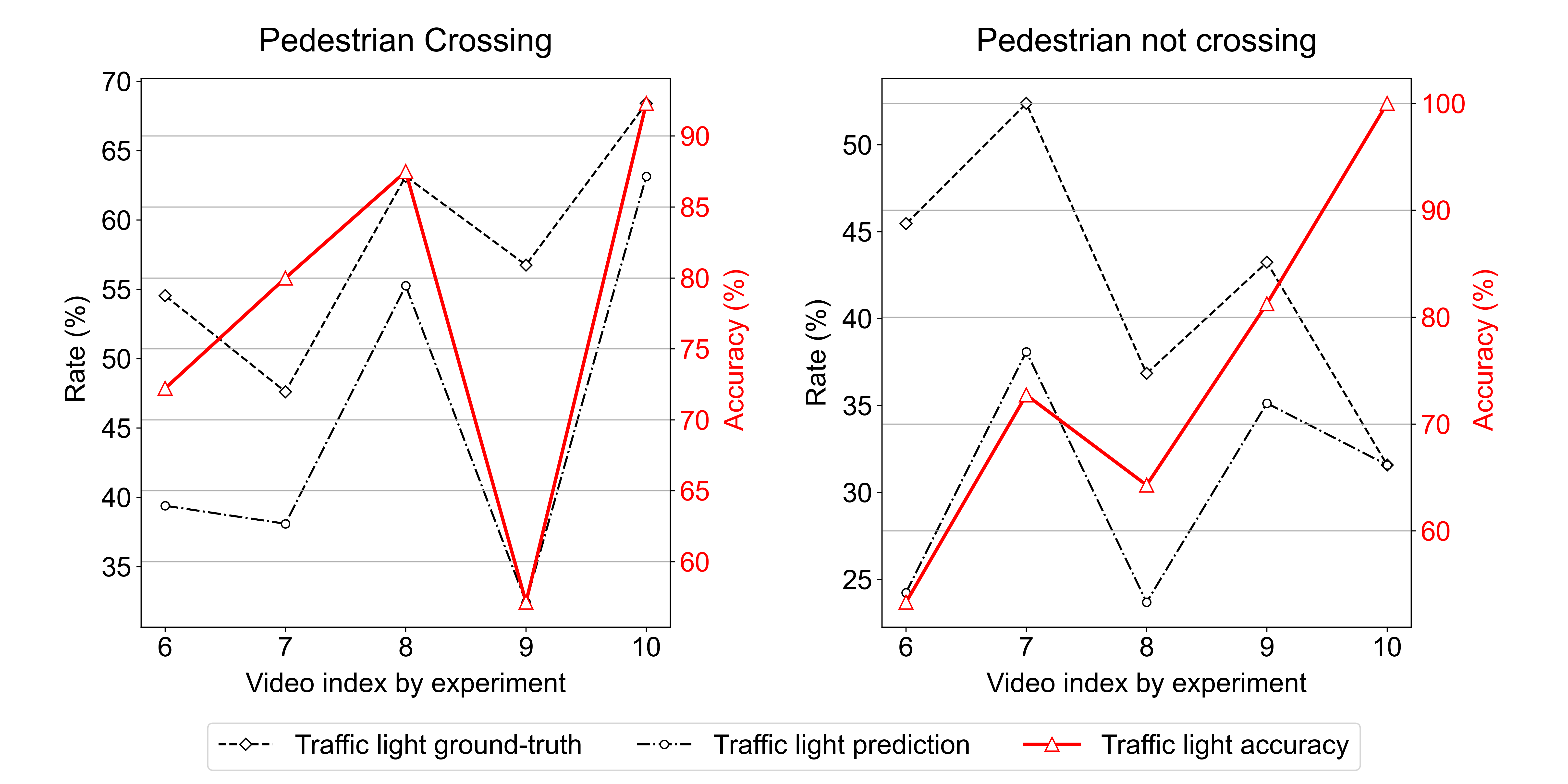}
		\caption{ }
	\end{subfigure}
	\caption{Graph depicts the accuracy and percentage of pedestrians in a) baseline and b) screen display c) traffic light condition, taking into account the ground-truth of the data and the results of the implemented prediction framework.}
	\label{fig:result}
\end{figure}

Figure~\ref{fig:result} shows the results. The graph depicts the real and estimated percentage of pedestrians who crossed or not, depending on the interface presented. For each interface, the ground-truth rate was estimated by a manual labelling process whereby it examined the videos to identify pedestrian actions. As it can be seen, if the ground-truth and estimation plots matched, a  100\% accuracy rate was obtained. The closer both curves are to each other, the higher the accuracy of the model.\\
In the case of crossing pedestrians, a minimum accuracy of 57.14\% and a maximum of 92.30\% was obtained in the prediction with the traffic light interface. In the case of non-crossing pedestrians we obtained a minimum value of 50\% under baseline conditions and a maximum value of 100\% in both baseline and light traffic conditions.

\begin{figure}
	\centering
	\includegraphics[width=0.48\textwidth]{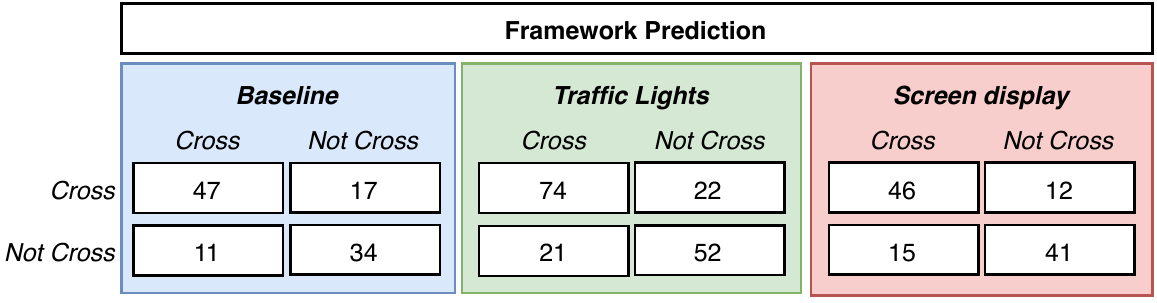}
	\caption{ Confusion matrix of the intention prediction framework regarding the different tests performed}
	\label{fig:decision}
\end{figure}

The number of false positives, false negatives, true positives, and true negatives is detailed in the confusion matrix in Figure~\ref{fig:decision}. The accuracy, precision, specificity and recall to measure the performance of the crossing intention prediction framework are depicted in Table~\ref{table:result}.

The intention prediction framework performance for crossing or no crossing behavior resulted in an average accuracy of 75\%, a precision of 78.04\% and a specificity of 71.35\%. The individual results of each test, showed a greater accuracy, specificity and ``cross'' recall in the case of screen display, and a greater precision and ``not cross'' recall in the baseline condition.

\begin{table}[]
	\centering
	%\raggedleft
	%\tiny
	\scriptsize
	\caption{Framework prediction performance depending on the tested scenario}
	\label{table:result}
	\begin{tabular}{|p{2cm}|p{1cm}|p{1cm}|p{1cm}|p{1cm}|}
		\hline
		& \textbf{Baseline} & \textbf{Traffic Lights} & \textbf{Screen Display} & \textbf{Average} \\
		\hline
		\textbf{Accuracy}         & 0.74      & 0.75              & 0.76            & 0.75            \\
		\hline
		\textbf{Precision}        & 0.81      & 0.78             & 0.75             & 0.78     \\
		\hline
		\textbf{Specificity}        & 0.67       & 0.70          & 0.77             & 0.71     \\
		\hline
		\textbf{Cross Recall}     & 0.73       & 0.77            & 0.79             & 0.76    \\
		\hline
		\textbf{Not Cross Recall} & 0.76     & 0.71             & 0.73             & 0.73    \\
		\hline
	\end{tabular}
\end{table}
%%%%%%%%%%%%%%%%%%%%%%%%%%%%%%%%%%%%%%%%%%%%%%%%%%%%%%%%%%%%%%%%%%
\section{Discussion}

Results from the performed field tests showed 75\% overall accuracy,  25\% of errors being due to the accumulated errors along each one of the stages of the prediction of pedestrian intentions. 

In many cases, they were based on the individual movement of the vehicle, as the prediction model takes into account the 2D trajectory of each pedestrian, but in many cases the vehicle made a slight turn that altered the estimated trajectory of the previously detected pedestrian. This affected the input of the  prediction model used.

In addition, there were cases in which the pose estimation was corrupted due to the density of nearby pedestrians or  unfavorable lighting such as sunset, which altered the image recorded by the stereo camera.

In the cases in which the framework obtained an accuracy of 100\%,  few pedestrians were interacting with the vehicle.

A further factor influencing the results was that the scenario in which pedestrians and vehicles shared the same road, both behaving differently than they would in a conventional public road. The proposed model needs to be fully optimized for this kind of environment, a topic for future research. 

%%%%%%%%%%%%%%%%%%%%%%%%%%%%%%%%%%%%%%%%%%%%%%%%%%%%%%%%%%%%%%%%%%
\section{Conclusion and Future Work}
\label{sec:conclusion}

To avoid potentially unsafe road situations, vehicular units and other road users need to take each other into account.

Particularly when driverless vehicles are involved, it is crucial that the intentions of all road users are understood so that the corresponding actions can be performed according to each case. 

To this end we developed a framework for estimating the crossing intention of pedestrians when they were exposed to an autonomous vehicle in a real world situation. The framework consisted of the integration of different open source algorithms that allowed us to address different individual aspects for the pedestrian crossing prediction, such as pedestrian detection and pose estimation, pedestrian distance estimation, tracking and intention prediction models.

The framework was deployed in a real vehicle and tested with three experimental cases that showed a variety of communication messages to pedestrians in a shared space scenario. 

Future work will aim at solving the errors obtained by integrating other data sources into the model, such as the distance between vehicle and the pedestrian, or the 3D pedestrian path.

%%%%%%%%%%%%%%%%%%%%%%%%%%%%%%%%%%%%%%%%%%%%%%%%%%%%%%%%%%%%%%%%%%
%\vfill
\color{black}
\section*{ACKNOWLEDGMENT}
This work was supported by the Austrian Ministry for Climate Action, Environment, Energy, Mobility, Innovation and Technology (BMK) Endowed Professorship for Sustainable Transport Logistics 4.0 and the Spanish Ministry of Economy, Industry and Competitiveness under TRA201563708-R and TRA2016-78886-C3-1-R Projects.
\color{black}
%%%%%%%%%%%%%%%%%%%%%%%%%%%%%%%%%%%%%%%%%%%%%%%%%%%%%%%%%%%%%%%%%%
%\addtolength{\textheight}{-12cm}
%\vspace{10mm}
\bibliographystyle{IEEEtran}
\bibliography{paper,test}
\end{document}